\documentclass[10pt,twocolumn,letterpaper]{article}

\usepackage{cvpr}              

\usepackage{graphicx}
\usepackage{amsmath}
\usepackage{amssymb}
\usepackage{booktabs}
\usepackage{pifont}

\usepackage{algorithm} 
\usepackage{algpseudocode} 
\usepackage{capt-of}
\usepackage{xcolor}

\usepackage[pagebackref,breaklinks,colorlinks]{hyperref}

\newcommand{\greencheck}{{\color{green}\checkmark}}
\newcommand{\redx}{{\color{red}\xmark}}

\usepackage[capitalize]{cleveref}
\crefname{section}{Sec.}{Secs.}
\Crefname{section}{Section}{Sections}
\Crefname{table}{Table}{Tables}
\crefname{table}{Tab.}{Tabs.}

\def\cvprPaperID{976} 
\def\confName{CVPR}
\def\confYear{2023}

\begin{document}

\title{HARP: Personalized Hand Reconstruction from a Monocular RGB Video}

\newcommand*{\affaddr}[1]{#1}
\newcommand*{\affmark}[1][*]{\textsuperscript{#1}}
\newcommand*{\email}[1]{\small{\texttt{#1}}}
\author{
Korrawe Karunratanakul \quad
Sergey Prokudin \quad
Otmar Hilliges \quad
Siyu Tang \\
\affaddr{ETH Z{\"u}rich}, Switzerland\\
\email{\{korrawe.karunratanakul,sergey.prokudin,otmar.hilliges,siyu.tang\}@inf.ethz.ch} \\
{\small\url{https://korrawe.github.io/harp-project/}}  \vspace{-0.5em}
}


\newcommand{\myparagraph}[1]{\vspace{0.5em}\noindent\textbf{#1}}
\newcommand{\ST}[1]{{\color{green}[ST: #1]}}
\newcommand{\KK}[1]{{\color{magenta}[KK: #1]}} 
\newcommand{\SP}[1]{{\color{blue}[Sergey: #1]}} 
\newcommand{\OH}[1]{{\color{blue}[OH: #1]}}
\newcommand{\oh}[1]{\OH{#1}}
\newcommand{\st}[1]{\ST{#1}}
\newcommand{\otmar}[1]{\OH{#1}}
\newcommand{\todo}[1]{{\color{RedOrange}\textbf{TODO}: #1}}
\newcommand{\V}[1]{\mathbf{#1}}
\newcommand{\R}[0]{\rm I\!R}
\newcommand{\E}[0]{\rm I\!E}
\newcommand{\loss}[0]{\mathcal{L}}

\newcommand{\methodname}{{HARP}}
\newcommand{\methodnamefull}{{HALO: Hand ArticuLated Occupancy~}}

\newcommand{\vaename}{{HALO-VAE~}}

\newcommand{\norm}[1]{\left\lVert#1\right\rVert}

\newcommand{\cmark}{\ding{51}}%
\newcommand{\xmark}{\ding{55}}%

\twocolumn[{%
\renewcommand\twocolumn[1][]{#1}%
\maketitle

\begin{center}
  \newcommand{\teaserwidth}{\textwidth}
  \centerline{\includegraphics[width=0.98\linewidth]{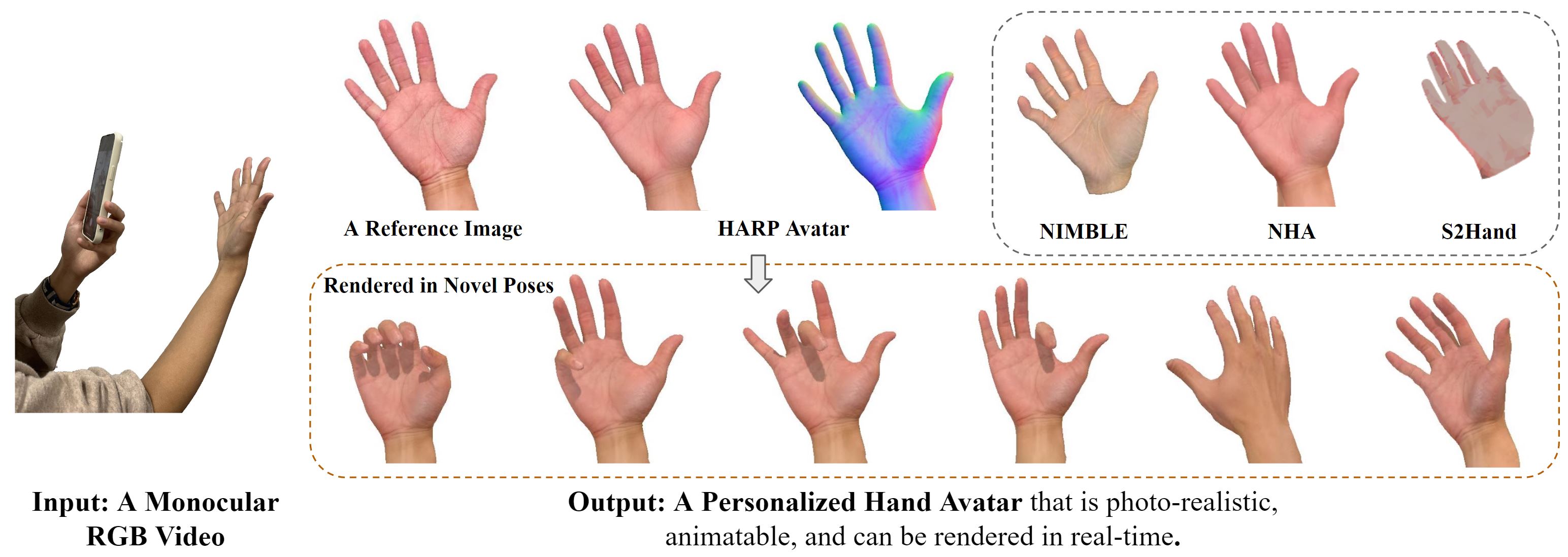}}
    \captionof{figure}{Given a short monocular video of a hand and a coarse hand pose and shape estimation for initialization, we reconstruct a photo-realistic hand avatar exhibiting faithful personalized appearance and geometry using standard explicit representations and a differentiable renderer {\it without any neural networks}. Compared to the baselines, our hand avatar demonstrates better-reconstructed geometry and appearance. The hand avatar can be used to render high-fidelity hand images in novel views and poses in real-time, which serves as a foundation for many AR/VR applications.}
    \label{fig:teaser}
\end{center}%
}]

\begin{abstract}
\vspace{-0.4cm}
We present \methodname~(HAnd Reconstruction and Personalization), a personalized hand avatar creation approach that takes a short monocular RGB video of a human hand as input and reconstructs a faithful hand avatar exhibiting a high-fidelity appearance and geometry.
In contrast to the major trend of neural implicit representations, \methodname~models a hand with a mesh-based parametric hand model, a vertex displacement map, a normal map, and an albedo without any neural components. 
The explicit nature of our representation enables a truly scalable, robust, and efficient approach to hand avatar creation as validated by our experiments. \methodname~is optimized via gradient descent from a short sequence captured by a hand-held mobile phone and can be directly used in AR/VR applications with real-time rendering capability.
To enable this, we carefully design and implement a shadow-aware differentiable rendering scheme that is robust to high degree articulations and self-shadowing regularly present in hand motions, as well as challenging lighting conditions.
It also generalizes to unseen poses and novel viewpoints, producing photo-realistic renderings of hand animations. 
Furthermore, the learned \methodname~representation can be used for improving 3D hand pose estimation quality in challenging viewpoints. 
The key advantages of \methodname~are validated by the in-depth analyses on appearance reconstruction, novel view and novel pose synthesis, and 3D hand pose refinement. 
It is an AR/VR-ready personalized hand representation that shows superior fidelity and scalability. 
\vspace{-0.5cm}
\end{abstract}


\section{Introduction}

Advancements in AR/VR devices are introducing a new reality in which the physical and digital worlds merge. The human hand is a crucial element for an intimate and interactive experience in these environments, serving as the primary interface between humans and the digital world.
Therefore, it is essential to capture, reconstruct, and animate life-like digital hands for AR and VR applications. Without this capability, the authenticity and practicality of AR/VR consumer products will always be limited.

Despite its importance, the research into hand avatar creation has so far been limited. Most works \cite{li2022nimble,HTML_eccv2020,chen2021s2hand} focus on creating an appearance space on top of a parametric hand model such as MANO \cite{MANO2017}. Such an appearance space provides a compact way to represent hand texture but is rather limited in expressivity to handle non-standard textures.
The recent LISA \cite{corona2022lisa} model has emerged as an alternative, using an implicit function to represent hand geometry and  texture color fields. Training a new identity in LISA, however, requires a multi-view capturing setup as well as a large amount of data and computing power.
In the nearby fields of face and body avatar creation, many works that leverage an implicit function \cite{grassal2021neural,Mihajlovic:CVPR:2022,LEAP:CVPR:21,MetaAvatar:NeurIPS:2021} or NeRF-based \cite{mildenhall2020nerf} volume rendering \cite{weng_humannerf_2022_cvpr,park2021nerfies,Mihajlovic:KeypointNeRF:ECCV2022} have also been recently explored.
The NeRF-based method such as HumanNeRF \cite{weng_humannerf_2022_cvpr} produces a convincing novel view synthesis but still shows blurry artifacts around highly articulated parts and cannot be easily exported to other applications.

We argue that democratizing hand avatar creation for AR/VR users requires a method that is {\it (1) accurate}: so that personalized hand appearance and geometry can be faithfully reconstructed; {\it (2) scalable}: allowing hand avatars to be obtained using a commodity camera; {\it  (3) robust}: capable of handling out-of-distribution appearance and self-shadows between fingers and palm; and {\it (4) efficient}: with real-time rendering capability.

To this end, we propose \methodname, a personalized hand reconstruction method that can create a faithful hand avatar from a short RGB video captured by a hand-held mobile phone. \methodname~leverages a parametric hand model, an explicit appearance, and a differentiable rasterizer and shader to reconstruct a hand avatar and environment lighting in an analysis-by-synthesis manner, 
{\it without any neural network component}. 
Our observation is that human hands are highly articulated. The appearance changes of observed hands in a captured sequence can be dramatic and largely attributed to articulations and light interaction. 
Learning neural representations, such as implicit texture fields \cite{corona2022lisa} or volume-based representations like NeRF \cite{su2021anerf}, is vulnerable to the over-fitting to a short monocular training sequence and can hardly generalize well to sophisticated and dexterous hand movements. 
By properly disentangling geometry, appearance, and self-shadow with explicit representations, 
\methodname~can significantly improve the reconstruction quality and generate life-like renderings on novel views and novel animations performing highly articulated motions. 
Furthermore, the nature of the explicit representation allows the results from HARP to be conveniently exported to standard graphics applications. 

In summary, the key advantages of \methodname~are:
(1) \methodname~is a simple personalized hand avatar creation method that reconstructs high-fidelity appearance and geometry using only a short monocular video. \methodname~demonstrates that an explicit representation with a differentiable rasterizer and shader is enough to obtain life-like hand avatars. 
(2) The hand avatar from HARP is controllable and compatible with standard rasterization graphics pipelines allowing for photo-realistic rendering in AR/VR applications.
(3) Moreover, \methodname~can be used to improve 3D hand pose estimation in challenging viewpoints.
We perform extensive experiments on the tasks of appearance reconstruction, novel-view-and-pose synthesis, and 3D hand poses refinement. 
Compared to existing approaches, \methodname~is more accurate, robust, and generalizable with superior scalability.

\section{Related Work}

\myparagraph{Hand Models.}
 Hand models are crucial for compactly representing a hand surface and using it in downstream applications. 
Many models rely on an explicit mesh surface \cite{Moon_2020_ECCV_DeepHandMesh,MANO2017,li21piano,xu2020ghum,li2022nimble}, while others implicitly represent the surface with neural networks \cite{corona2022lisa,karunratanakul2021halo,karunratanakul2020grasping}.
The widely used MANO model \cite{MANO2017} represents a hand with pose and shape vectors, generating hand meshes using a PCA model and linear blend skinning.
While parametric models like MANO can handle various hand geometries, their expressivity is limited by the learned parameter spaces. An alternative approach is to store vertex locations directly to broaden geometry representation \cite{ge20193d,moon2020i2l,lin2021metro,Smith2020ConstrainingDH}, but this prevents re-animation. Conversely, a limited amount of work has been explored for hand appearance modeling.
Notable works include HTML \cite{HTML_eccv2020}, a linear appearance model for inferring the UV texture on top of MANO, and
NIMBLE \cite{li2022nimble}, an extension of the hand skeleton model \cite{li21piano}, which can infer the surface appearance, including an albedo, specular, and normal map.
Despite their realistic texture, these model suffers from their linear nature of appearance space and limited appearance data, making them unsuitable for adapting to a novel identity.
In this work, we enhance the parametric model with personalized geometry adjustment, explicit albedo, and normal maps, enabling animatable personalized geometry and stronger texture representation than PCA-based textures.

\myparagraph{Geometry Reconstruction.}
To build a hand avatar, one must first derive the hand geometry from the input images which has been a long-studied problem \cite{BTGG12,mueller_siggraph2019,rhoi2020fitting,rudnev2021eventhands,tang2021towards,chen2021i2uv,ng2021body2hands,zhang2021single}.
To estimate the hand surface, these methods generally leverage the statistical prior in the MANO model by predicting its pose and shape parameters.
One advantage of such a method is that it could prevent geometry from collapsing by encouraging the hand shape to be close to the mean shape \cite{hasson2019obman,karunratanakul2020grasping}.
To overcome this limited expressiveness of the MANO shape space, a convolution neural network (CNN) can be used to directly estimate the vertex locations of the mesh topology \cite{ge20193d,Choi_2020_ECCV_Pose2Mesh,kolotouros2019cmr,Kulon_2020_CVPR,moon2020i2l}.
Lin \etal \cite{lin2021metro} replace the convolution operation with a Transformer model \cite{vaswani2017attention} and achieve state-of-the-art results on various datasets.
Alternatively, the hand surface can be represented by an implicit function \cite{karunratanakul2021halo,karunratanakul2020grasping,corona2022lisa}. With HALO \cite{karunratanakul2021halo}, the hand geometry can be estimated by any key point estimation methods \cite{mueller2018ganerated,garcia2018first,iqbal2018hand,cai2018weakly,moon2018v2v,tekin2019ho,yang2019disentangling,doosti2020hope,spurr2020eccv,moon2020interhand,zimmermann2017learning}. 
Given ground truth hand poses in multi-view images, differentiable ray tracing can also be used to refine the annotations \cite{karvounas2021multi}.
In this work, we leverage the prediction by METRO \cite{lin2021metro} as initialization to refine a personalized geometry.

\myparagraph{Appearance Representations.}
Numerous methods have been proposed to learn and estimate appearance from images or videos for bodies with clothing \cite{prokudin2021smplpix,Siarohin_2019_NeurIPS,peng2021neuralbody,HVTR:3DV2022,li2022tava,noguchi2021neural_art} and faces \cite{buehler2021varitex,DECA_Siggraph2021,grassal2021neural,gafni2021dynamic,kim2018deep,park2021nerfies}, including those that use NeRF \cite{mildenhall2020nerf} to implicitly represent appearance \cite{park2021nerfies,park2021hypernerf}.
To model a human, pose information can also be used as conditions to transform the radiance field of a human avatar \cite{2021narf,weng_humannerf_2022_cvpr,su2021anerf,chen2022relighting,xu2021h}.
However, these models are not suitable for hand appearance due to the high degree of articulation. Furthermore, mesh extraction is required to make these methods compatible with traditional graphics applications. 
In contrast, less research has been done on hand appearance representation \cite{seeber2021realistichands}.
While LISA \cite{corona2022lisa} and HandAvatar \cite{handavatar2022Chen} both learn implicit color fields together with implicit surface representations, they require a large number of images to train a subject-specific shape and appearance.
Given an image, S2hand \cite{chen2021s2hand} employs an MLP to estimate the vertex colors and lighting together with MANO parameters.
Concurrently, Wang \etal proposed SunStage \cite{wang2022sunstage}, an outdoor face reconstruction method that shares similarities to our method, in particular, the use of explicit mesh representation and differentiable rendering.
%
In this work, we present an explicit hand appearance model and the optimization pipeline that can capture detailed hand texture while also taking lighting and shadowing into account.

\begin{figure*}[t]
    \centering
    \includegraphics[width=1.0\linewidth]{{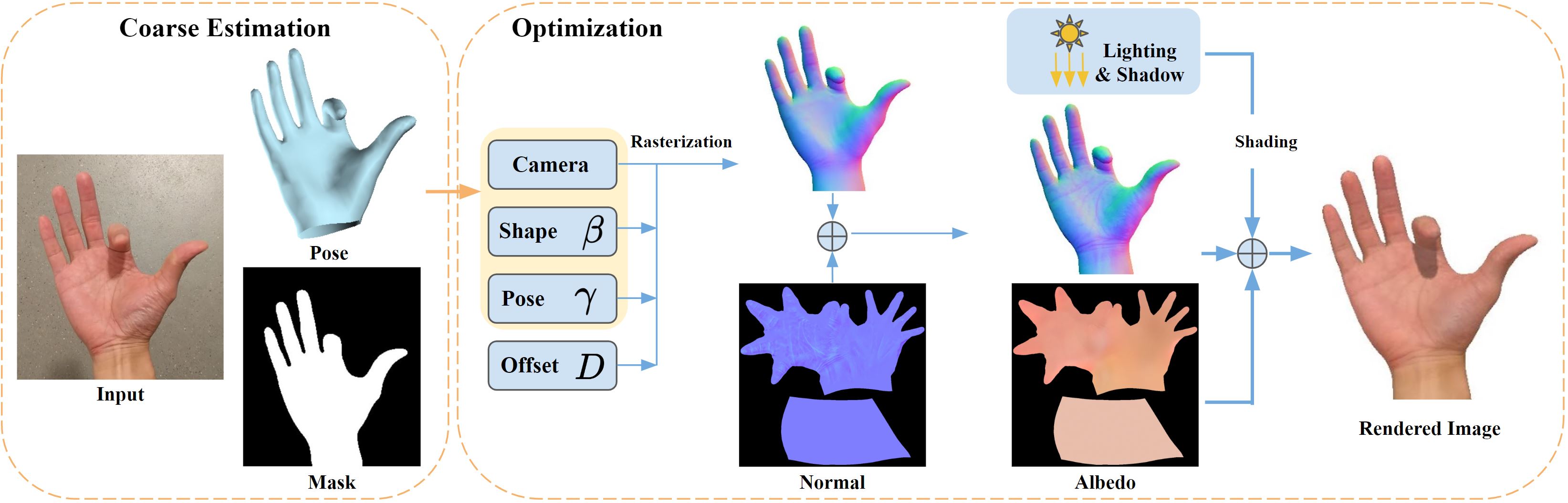}}
    \caption{\textbf{Method Overview.} Given a short monocular RGB video of a hand, our hand avatar creation method includes two steps: (1) coarse hand pose and shape estimation for initialization; (2) an optimization framework to reconstruct the {\it personalized} hand geometry and appearance. The hand geometry is first rasterized and combined with a normal map. Then, the shader combines the albedo, geometry, and lighting to render the personalized hand. The optimization solves the hand and scene parameters using only the input images.}
    \label{fig:pipeline}
    \vspace{-1em}
\end{figure*} 

\section{Method}
\textbf{Overview.} The overview of \methodname~is illustrated in Fig.~\ref{fig:pipeline}. 
Given a short monocular RGB video of a hand consisting of $N$ frames, we reconstruct a realistic hand avatar with {\it personalized} shape and texture. Specifically, our method outputs a triangle mesh $M$, containing vertices $V$ and faces $F$, and its UV texture, which is decomposed into albedo and normal maps.
We optimize the personalized hand mesh $M$, the albedo, and the normal map using an analysis-by-synthesis approach by comparing the input images to the images of our reconstruction $M$ rendered using a differentiable rendering framework.
\\
Our approach focuses on \textit{efficiency}, \textit{robustness}, and \textit{exportability}, while also maintaining a \textit{high-quality hand appearance}.
Given the focus, we employ 
(1) explicit representations (mesh, normal map, and albedo) which can be easily exported to any graphics application; (2) direct optimization of the explicit appearance without relying on a learned appearance space, such that we do not require pre-training nor a large number of training images. 
(3) the explicit and efficient rasterizing and shading which provides a good balance between rendering quality and computation cost.

\subsection{Hand Representation} \label{sec:hand_rep}
\myparagraph{Template Model.}
Our hand template model is built upon the MANO~\cite{MANO2017} model, which we extend to a higher mesh resolution and allow surface vertex deformation from the template.
Concretely, we perform a linear subdivision on the MANO template to increase the number of vertices from 778 to 3093, which in turn allows our template to capture more surface details.
The subdivision process is differentiable, thus enabling gradient propagation back to the MANO pose parameter $\gamma$ and shape parameter $\beta$. 
Additionally, the MANO hand is truncated at the wrist, which does not reflect the reality where hands are attached to the arms and the foreground mask does not separate the wrist. Therefore, we built another template from SMPLX \cite{SMPL_X2019} by truncating the arm at the elbow to facilitate the hand-and-arm fitting depending on the available mask.
Their interactions with the rest of the system remain identical.

\myparagraph{Geometry Refinement.}
To utilize the higher mesh resolution for finer geometry details, we allow each posed vertex to additionally deform based on a personalized vertex displacement $D$ along the vertex normal.
The posed hand vertex locations without arm could be obtained with:
{\small
\begin{eqnarray} 
    & V =  \mathcal{S}(\mathcal{M}(\gamma, \beta)) + D, \\
    & \mathcal{S}: \mathbb{R}^{778 \times 3} \to \mathbb{R}^{3093 \times 3},
\end{eqnarray}
}%
where $\mathcal{M}$ is the MANO function which takes pose $\gamma$ and shape $\beta$ as inputs and returns posed mesh vertex locations, $\mathcal{S}$ is the mesh subdivision function.
The vertex displacement vector $D$ is subject-specific, which we obtain by optimizing it along with other parameters.

\myparagraph{Texture.} \label{sec:texture_model}
We model the hand skin as a Lambertian surface with an albedo map $a$, which is defined per subject in a UV space.
Additionally, to add high-frequency details without upsampling the mesh, we utilize a UV-space normal map which can be combined with the surface normal $\hat{N}$ when computing the illumination at the 3D surface point $x$.

\subsection{Shadow-aware Differentiable Rendering} \label{sec:lighting}
There are multiple options for differentiable rendering such as a differentiable path tracer \cite{NimierDavidVicini2019Mitsuba2}, differentiable rasterizer \cite{liu2019softras}, neural renderer \cite{kato2018neural} or the recently developed NeRF-based volume rendering \cite{mildenhall2020nerf}.
The path tracer is known for its realistic rendering and explicit factorization of material properties. However, this comes at a high computation cost and has not been shown to work with highly articulated objects \cite{hasselgren2022nvdiffrecmc,NimierDavidVicini2019Mitsuba2}.
NeRF-based methods \cite{su2021anerf,weng_humannerf_2022_cvpr,zhang2021nerfactor} and implicit texture-based methods \cite{corona2022lisa} are often computationally expensive, and as they leverage MLPs to approximate the light interaction with the articulated parts; the result can be blurry and have self-shadows baked into the texture representation.

In this work, we demonstrate that, to create a realistic hand avatar from a short video sequence, it is not necessary to rely on neural implicit representations, such as neural volume rendering.
Standard explicit geometry and appearance representations together with a carefully implemented differentiable rendering scheme are able to provide the right balance between quality, speed, and compatibility with other graphics applications.

To render the hand in the camera view, we first utilize the differentiable rasterizer \cite{ravi2020pytorch3d} to determine the visible surfaces from the camera views.
We then use the Phong model \cite{phong1975illumination} without the specular component to compute the illumination, i.e. color, at the surface point $x$:
 \begin{eqnarray}
    I_x &= k_a i_a + \sum\limits_{m \in \text{lights}} k_d 
i_{m,d} (\hat{L}_m \cdot \hat{N})
 \end{eqnarray}
where $k_a$ is an ambient reflection constant, $k_d$ is a diffuse reflection constant, $i_a$ and $i_{m,d}$ are intensities of the light sources for diffuse surface, $\hat{N}$ is the normal at point $x$, and $\hat{L}_m$ is a ray from the surface point toward each light source.
We observe that having just one dominant light source ($|m|=1$) provides a good balance between computation cost and the rendering quality, under the assumption that the hard shadow is usually produced by the closest light source when indoors and by the sun outdoors.

\myparagraph{Self-shadowing.}
Notably, the Phong model does not account for self-shadowing which often occurs when a finger casts a shadow onto other fingers and the palm.
To accommodate this scenario, we add the visibility term $V(x,m)$ at $x$ with respect to the light $m$ to the diffuse component, making it $k_d(\hat{L}_m \cdot \hat{N})V(x,m)i_{m,d}$.
We compute the visibility $V$ by performing a two-step rasterization of the mesh.
We integrate the classic idea of shadow mapping in computer graphics \cite{shadow_map} into our differentiable pipeline. \\
The step-by-step computation is shown in Alg~\ref{alg:vis}.
The $z$-buffer $Z_m$ is a depth image when seen from the camera $cam_m$.
If the light position is more than 1 m away from the hand, we project it to 1 m distance.
The Sigmoid function is used to ensure smooth gradients, with a bias term $b=0.005$ and scale $s=1000$.
To produce a softer shadow, we use percentage-closer filtering \cite{reeves1987rendering}, which averages the visibility values of the nearby points.
These visibility values then allow us to integrate the self-shadow into the pipeline.
We note that our self-shadowing implementation is differentiable with respect to both geometry and appearance.
Our implementation is compatible with the pyTorch3D \cite{ravi2020pytorch3d} package, and will be made publicly available.

\begin{algorithm}
\caption{Visibility $V(x,m)$ from light $m$} 
\small
\begin{algorithmic}[1]
\State Place virtual camera ${cam}_m$ at light $m$ pointing at hand
\State Compute $z$-buffer ${Z_m}(\cdot)$ from $cam_m$
\State Get 3D points $\{X_{hit}\}$ seen from the actual camera
\For{$x \in X_{hit}$}
    \State Transform $x$ to ${cam}_m$ coordinate 
    \Statex \hspace{\algorithmicindent} $x^m = T_m(x)$ 
    \State Get 2D pixel coordinate of $x^m$
    \Statex \hspace{\algorithmicindent} $x^{2d} = \pi(x^m)$
    \State Compute distance to the light
    \Statex \hspace{\algorithmicindent} $d_{m \rightarrow x} = ||x-m||$
    \State $V(x,m) = {Sigmoid}(s( Z_m(x^{2d}) - d_{m \rightarrow x} + b))$
\EndFor
\end{algorithmic}
\label{alg:vis}
\end{algorithm}

\subsection{Optimization}
To find the parameters that describe the personalized hand, we optimize the parameters using short RGB videos. The optimization objective is to minimize the difference between the input and the rendered hand images using the proposed differentiable rendering pipeline.
For each subject, we optimize for a joint objective consisting of a geometry alignment term $E_{geo}$ and an appearance term $E_{app}$: 
 \begin{eqnarray}
    E &= E_{geo} + E_{app},
 \end{eqnarray}
where $E_{geo}$ focuses on mesh configuration and $E_{app}$ encourages the same appearance as in the input images. 

\myparagraph{Geometry Objective.}
For the geometry, the goal is to match the rendered silhouette with the hand mask while also satisfying 3D mesh constraints.
The geometry objective is defined independently from the appearance as follows:
 \begin{eqnarray}
    E_{geo} &= w_{sil} \cdot E_{sil} + E_{reg},
 \end{eqnarray}
where $E_{sil}$ is the silhouette difference term, $E_{reg}$ is the mesh regularization term, and $w$ are the weights.

The silhouette difference term is the $l_1$-difference between the hand mask and the rendered silhouette $S_{render}$:
 \begin{eqnarray}
E_{sil} = |S_{in} - S_{render}|,
 \end{eqnarray}
where $S_{in} \in \{0,1\}^{H \times W}$ is an input binary hand mask that can be obtained from off-the-shelf segmentation tool \cite{unscreen}.

To prevent the optimized mesh from collapsing, we employ a combination of 3D mesh regularizations defined as:
 {\small
 \begin{eqnarray}
    E_{reg} &= E_{init} + E_{verts} + E_{lap} + E_{norm}  + E_{arap},
 \end{eqnarray}
 }
where each term is accompanied by its weight.

The term $E_{init}$ penalizes key points deviation from the initial pose with $l_1$ distance. The vertex offset regularization $E_{verts}$ controls the deviation from the MANO mesh to be small using an $l_2$-norm: $E_{verts} = \norm{D}^2$.

The mesh vertices $V$ are regularized by the Laplacian mesh regularizer \cite{desbrun1999implicit} $E_{lap}$ and the normal consistency regularizer $E_{norm}$ defined on the posed mesh.

The term $E_{arap}$ is the as-rigid-as-possible term \cite{ARAP_modeling:2007} that encourages the 3D mesh to be more rigid and distributes the changes in length among multiple edges.
The edge length difference is defined with respect to the MANO template as:
 \begin{eqnarray}
    E_{arap} &= \sum_v^V \sum_{u \in \mathcal{N}(v)} \norm{ \norm{v_t - u_t} - \norm{ v^* - u^* } }^2 ,
 \end{eqnarray}
where $\mathcal{N}(v)$ are the adjacent vertices of $v$, $v_t$ is a vertex from frame $t$, and $v^*$ is a vertex from the MANO template.

\myparagraph{Appearance Objective.}
The appearance term $E_{app}$ measures the similarity between the input and the rendered image. Note that as the texture is mapped to the triangle mesh for rendering, the appearance term is also affected by the geometry change.
We define the appearance term $E_{app}$ as:
 \begin{eqnarray}
    E_{app} &= w_{photo} \cdot E_{photo} + w_{vgg} \cdot E_{vgg} + E_{app\_reg}
 \end{eqnarray}
where $E_{photo}$ is a per-pixel $l_1$ color difference between the input images and the predicted images, $E_{vgg}$ is the VGG loss \cite{ledig2017photo} that captures the perceptual difference between the two images by comparing the features extracted using the VGG model \cite{simonyan2014vgg}, and $E_{app\_reg}$ is a regularization term that encourages both albedo and normal map to be locally smooth \cite{zhang2021nerfactor} (details in the Appendix).

\myparagraph{Initialization.}
For optimization, we initialize the hand parameters with per-frame predictions from the hand pose estimator METRO \cite{lin2021metro}.
As METRO directly predicts the mesh coordinate without using the MANO pose and shape space, we obtain the equivalent MANO pose $\gamma$ and shape $\beta$ parameters by minimizing the $l_2$-distances between corresponding vertices from the prediction and the MANO mesh.

\myparagraph{Optimization.}
In summary, we optimize: \textbf{hand geometry parameters}: (1)~$\beta$, the global MANO shape parameter, (2)~$D$, the per-vertex displacement, (3)~$\gamma$, the per-frame MANO pose parameter, including (4)~the per-frame translation, \textbf{global appearance parameters}: (5)~$a$, the UV-space albedo, and (6) the UV-space normal map, \textbf{lighting parameters}: (7)~$x_{lights}$, the light positions, (8)~$k$, the global reflection constant.
Fig.~\ref{fig:pipeline} shows the overview of our optimization process. More details are in Appendix.

\section{Experiments}

\begin{figure}
    \centering
    \includegraphics[width=\linewidth]{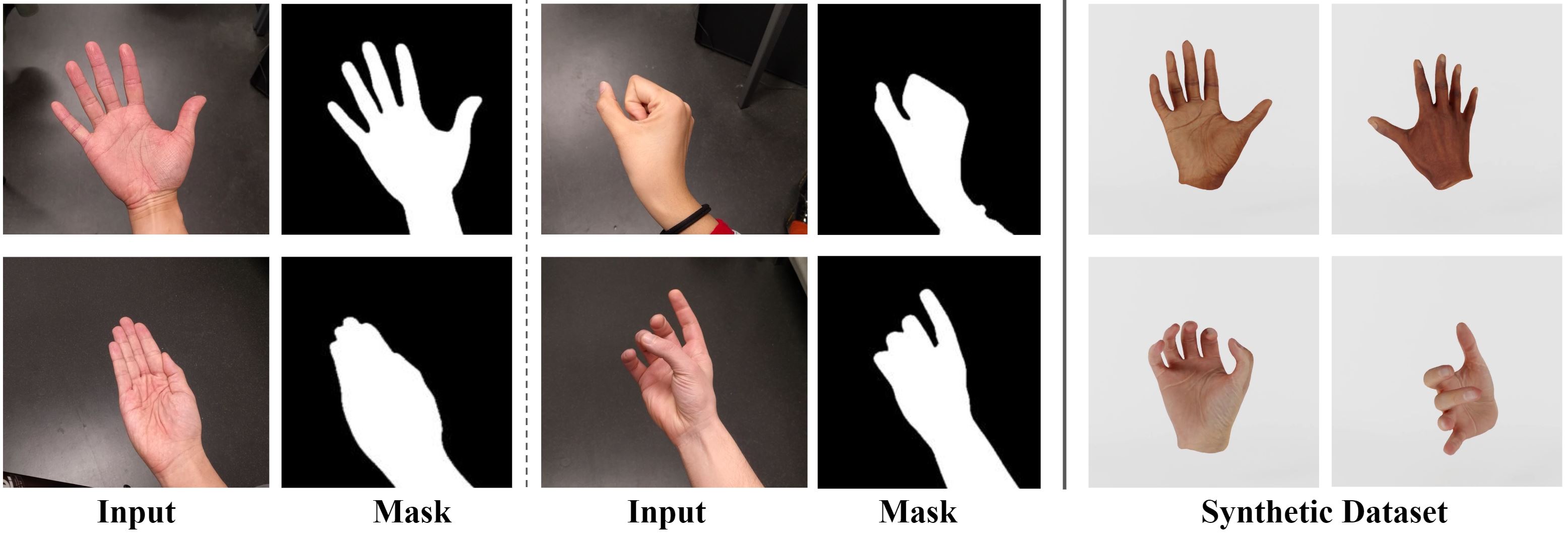}
    \caption{\textbf{Sample images.} (left) Each subject in the hand appearance dataset. (right) The synthetic dataset. The hand segmentation is obtained using an off-the-shelf segmentation tool \cite{unscreen}.}
    \label{fig:sample_data}
    \vspace{-1em}
\end{figure}

\myparagraph{Datasets.}
There is a rich literature on datasets for hand pose estimation and geometry reconstruction \cite{moon2020interhand,zimmermann21hanco,chao2021dexycb
}, but less on hand appearance reconstruction. 
For our personalized hand avatar reconstruction task, there is no suitable dataset captured in the out-of-lab environment with monocular commodity equipment. 
Therefore, apart from evaluating on the existing InterHand2.6M \cite{moon2020interhand} dataset, we create our own datasets for hand avatar creation, including a hand appearance dataset and a synthetic dataset (Fig. \ref{fig:sample_data}).

{\it Hand Appearance Dataset.} 
To simulate less-constrained capture settings similar to what the end-users of AR/VR applications typically have, all of our videos are captured by holding a phone camera pointing at a right hand in normal office lighting (Fig.\ref{fig:teaser}).
Our hand appearance dataset consists of three parts:
\textbf{(1) Multi-subject single-view hand sequences.}
The captures contain four subjects, three male subjects and one female subject, in motions ranging from flipping the hand to more complex interactions between fingers.
In total, there are 750 training and 600 testing frames for each subject.
\textbf{(2) Out-of-distribution hand appearance.}
We captured 2 additional subjects with tattoos for testing the out-of-distribution appearance, each containing 4 sequences. Otherwise, the setting is the same as the first part.
\textbf{(3) Lighting and shadow variation.}
We selected a subject from the previous part to capture 6 sequences of simple hand motions under varying directions of a single dominant light source. The shadow is highly pronounced in this portion.
For all parts, we ensure that both sides of the hand are visible. It is still possible that parts that are usually occluded, such as areas between fingers, are not visible.
We obtain a binary hand mask for each frame using an online segmentation tool \cite{unscreen} that considers the hand and the visible part of the arm as foreground.

{\it Interhand2.6M Dataset.} \cite{moon2020interhand}
The data is captured in a capturing dome with controlled lighting which is more restrictive than our primary goal of casually captured video.
The foreground masks are obtained using RVM \cite{rvm_seg}, which are sometimes noisy.
We select a single-view sequence of length 500 frames from the test set where most of the surface is visible for appearance evaluation.

{\it Synthetic Dataset.}
As the 3D annotations of real datasets often contain fitting error (reported at 2-3mm for InterHand2.6M \cite{moon2020interhand}), we opt to use a synthetic dataset with perfect ground truth to evaluate hand pose estimation task.
We rendered images of two subjects, each with two sequences, using a ray tracing engine Cycles in Blender \cite{blender}.
More details about the datasets can be found in the Appendix.

\myparagraph{Baselines.}
We summarize the overview of the available baselines for hand avatar creation from RGB images in Tab.~\ref{table:model_overview}.
Our main advantages compared to the baselines are: (i) by relying on explicit UV and normal maps, our output is directly compatible with standard graphics pipelines; (ii) we can represent non-standard hand textures which are often not captured well by the models that rely on a PCA space for appearance (Sec~\ref{sec:app_ood}); (iii) HARP is more efficient in term of optimization time and memory than MLP-based methods due to the lower number of parameters.

For the methods which are a variation of a hand model, namely S2Hand \cite{chen2021s2hand}, HTML \cite{HTML_eccv2020}, and NIMBLE \cite{li2022nimble}, the appearance is often predicted by a regressor, which can be inaccurate. Nevertheless, such models also allow test time optimization to refine the appearance according to the input images. For a fair comparison, we \textbf{optimize} the hand appearances in all of these baselines at test time.

We provide details on how we adapt each work for the hand avatar creation in the Appendix.
For the methods that require pose and shape initialization--all except S2Hand--we fit the hand model to the vertex predictions by METRO\cite{lin2021metro}.
The comparison to LISA \cite{corona2022lisa} is omitted as there is no released code, model, or result that we could compare to. 
Nevertheless, our results on InterHand2.6M (Fig.~\ref{fig:other_dataset}) show a superior qualitative appearance than those presented in \cite{corona2022lisa}.

\begin{table} 
\footnotesize
\centering
\begin{tabular}{lccc}
\toprule
 & Appearance Rep. & Out-of-dist.  & Compatibility \\ \hline

\multicolumn{2}{l}{w/ shared appearance space} \\
{S2Hand \cite{chen2021s2hand}} & vertex color &  \redx &  \greencheck  \\ 
{HTML \cite{HTML_eccv2020}} &  UV map &  \redx &  \greencheck   \\ 
{NIMBLE} \cite{li2022nimble} & UV+spec+normal &  \redx &  \greencheck   \\ \hline

\multicolumn{2}{l}{w/o shared appearance space} \\
{LISA} \cite{corona2022lisa} &  3D Implicit &  \greencheck &  \greencheck$^1$   \\ 
{NHA* \cite{grassal2021neural}} &  2D Implicit &  \greencheck &  \redx$^2$  \\ \hline

{HARP} &  UV + normal &  \greencheck &  \greencheck \\ 
\bottomrule
\end{tabular}
\caption{\textbf{Overview of hand appearance models}. Appearance Rep indicates the method used to represent texture information in the model. Out-of-dist refers to its ability to represent arbitrary non-standard hand appearance, including hands with scars or tattoos. Compatibility indicates if the method could export the 3D model with appearance for use in other applications.
*For Neural Head Avatar (NHA), we adapt the official code to hand domain. To make it compatible with other methods, we dropped the face-specific components from the model, including face landmarks, segmentation, and normals.
$^1$Can be extracted as a mesh by Marching Cubes.
$^2$Geometry is compatible but colors cannot be extracted.
}
\vspace{-1em}
\label{table:model_overview}
\end{table}

\myparagraph{Evaluation Metrics.}
For appearance evaluation, we report metrics that focus on the rendered image quality including the pixel-wise L1, the silhouette intersection-over-union (IoU), the learned perceptual image patch similarity (LPIPS) \cite{zhang2018unreasonable}, and the multi-scale structural similarity metric (MS-SSIM) \cite{wang2003multiscale}. The rendered image with white background is compared to the masked input.
Importantly, the pixel-wise L1 difference, LPIPS, and the intersection-over-union evaluation are not directly comparable between methods that produce different hand geometry due to the missing appearance of the truncated wrist in MANO and NIMBLE. Therefore they should be considered as references rather than direct comparisons.
For pose evaluation, we report Procrustes-aligned vertex error (PA-MPVPE) in mm compared to the MANO ground truth.

\subsection{Hand Appearance Reconstruction} \label{sec:app_recon}
\begin{figure}
    \centering
    \includegraphics[width=1.0\linewidth]{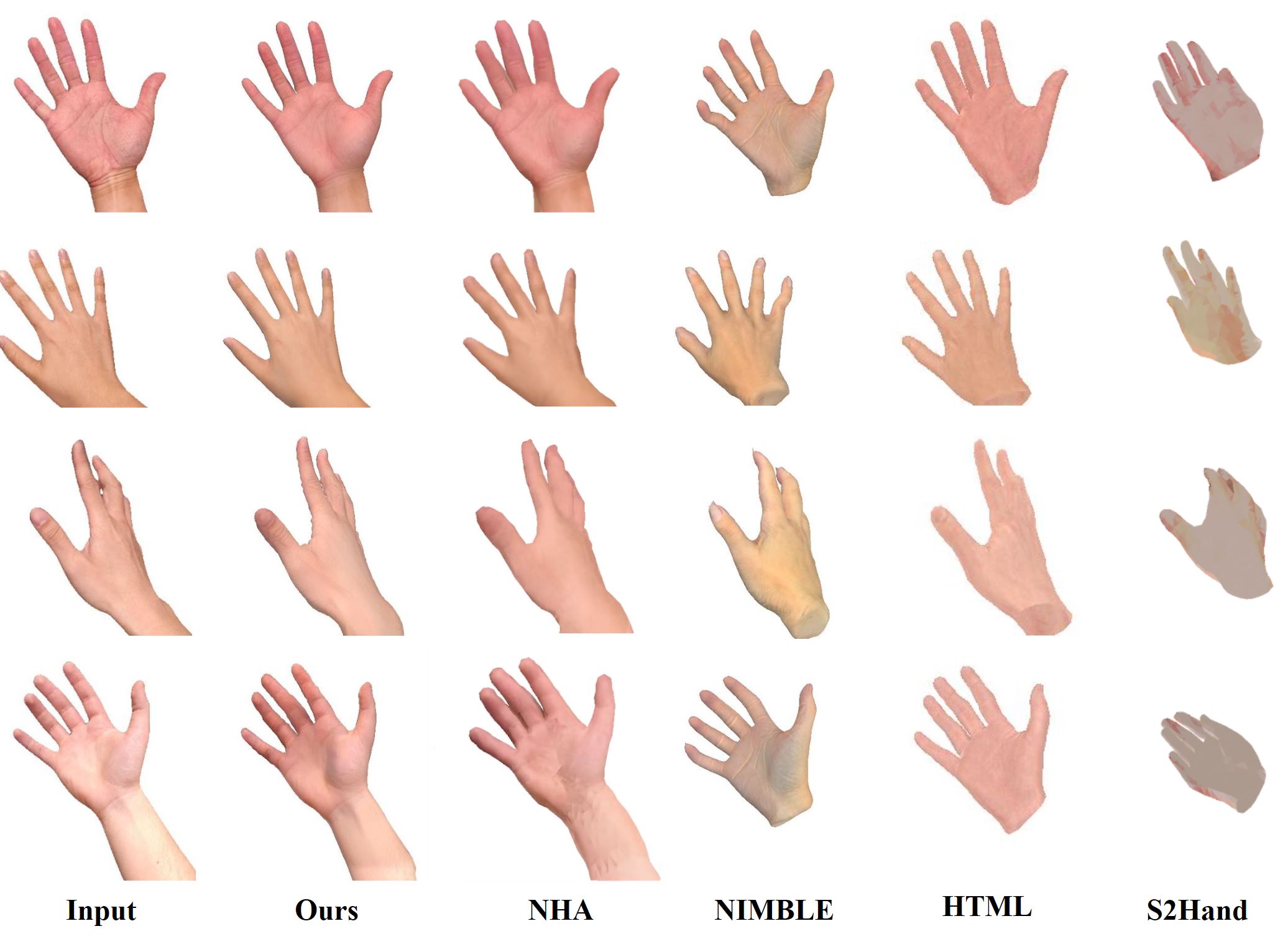}
    \caption{Qualitative comparison to the baselines on our captured sequences. Zoom in for details.}
    \label{fig:result_texture_recon}
    \vspace{-1em}
\end{figure}

To demonstrate the realism and robustness of HARP, we first evaluate its ability to reconstruct and re-render hand appearance from RGB sequences.
We use the first part of our appearance dataset which reflects a more common environment for hand-related applications, e.g., personalized hand reconstruction and rendering for AR/VR. 
We show the qualitative texture and geometry in Fig.~\ref{fig:result_texture_recon} and the quantitative evaluation in Tab.~\ref{table:result_texture_recon} averaged over all subjects.
The results suggest that HARP can faithfully reconstruct hand appearance with much higher details than the baselines.
In addition, we demonstrate the robustness of HARP with appearance evaluation on the InterHand2.6M sequence in Tab.~\ref{table:result_texture_recon_interhand} and avatar reconstruction on various datasets in Fig.~\ref{fig:other_dataset}.

\begin{table} 
\footnotesize
\centering
\begin{tabular}{lcccc}
\toprule
 & IoU $\uparrow$~ & L1 $\downarrow$~ & LPIPS $\downarrow$~ & MS-SSIM $\uparrow$~ \\ \hline

\multicolumn{1}{l|}{METRO \cite{lin2021metro}} & 0.651 & - & - & - \\ 
\multicolumn{1}{l|}{S2Hand \cite{chen2021s2hand}}  & 0.430 &  0.058 & 0.270  & 0.595  \\ 
\multicolumn{1}{l|}{HTML \cite{HTML_eccv2020}}  & 0.778 & 0.033  & 0.136  & 0.791  \\ 
\multicolumn{1}{l|}{NHA \cite{grassal2021neural}}  & 0.860 &  0.025 & 0.114  & 0.878  \\ 
\multicolumn{1}{l|}{NIMBLE \cite{li2022nimble}}  & 0.641 &  0.048 & 0.204  & 0.691  \\ 
\hline
\multicolumn{1}{l|}{HARP} & \textbf{0.929} & \textbf{0.018} & \textbf{0.071} & \textbf{0.902} \\ 
\bottomrule
\end{tabular}
\caption{{Quantitative evaluation of the appearance reconstruction task on the train split of our captured sequences}.}
\label{table:result_texture_recon}
\end{table}

\begin{table} 
\footnotesize
\centering
\begin{tabular}{lcccc}
\toprule
 & IoU $\uparrow$~ & L1 $\downarrow$~ & LPIPS $\downarrow$~ & MS-SSIM $\uparrow$~ \\ \hline

\multicolumn{1}{l|}{METRO \cite{lin2021metro}} & 0.561 & - & - & - \\ 
\multicolumn{1}{l|}{HTML \cite{HTML_eccv2020}}  & 0.571 & 0.091  & 0.203  & 0.822  \\ 
\multicolumn{1}{l|}{NHA \cite{grassal2021neural}}  & 0.651 &  0.084 & 0.229  & 0.819  \\ 
\multicolumn{1}{l|}{NIMBLE \cite{li2022nimble}}   & 0.621 &  0.083 & 0.229  & 0.775 \\ 
\hline
\multicolumn{1}{l|}{HARP} & \textbf{0.779} & \textbf{0.051} & \textbf{0.173} & \textbf{0.876} \\ 
\bottomrule
\end{tabular}
\caption{Appearance reconstruction results on InterHand2.6M.}
\label{table:result_texture_recon_interhand}
\end{table}

\begin{figure}
    \centering
    \includegraphics[width=1.0\linewidth]{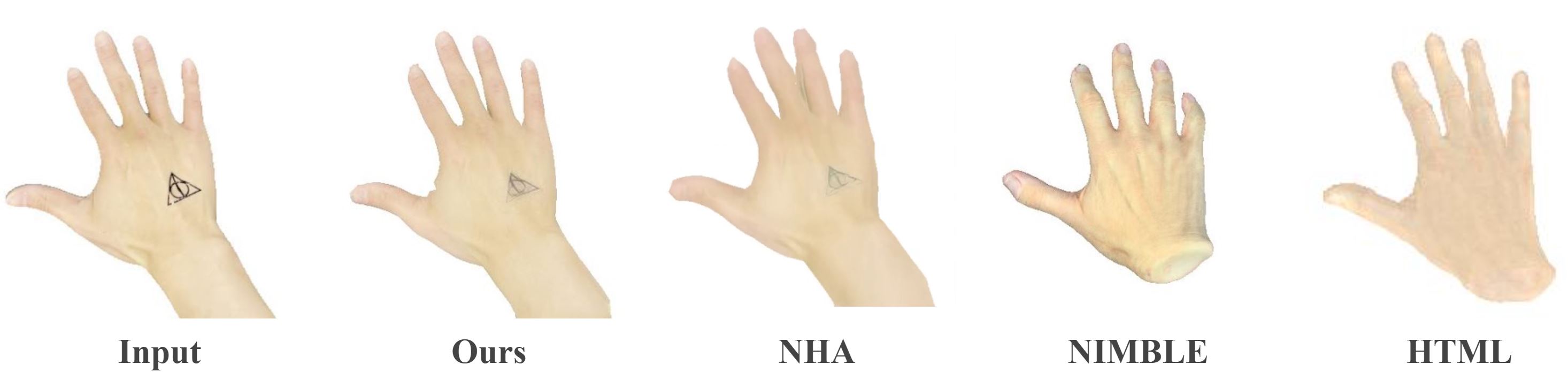}
    \caption{Qualitative comparison between our method and baselines on hands with out-of-distribution appearance.}
    \label{fig:tattoo}
\end{figure}

\begin{figure*}[h]
    \centering
    \includegraphics[width=0.95\linewidth]{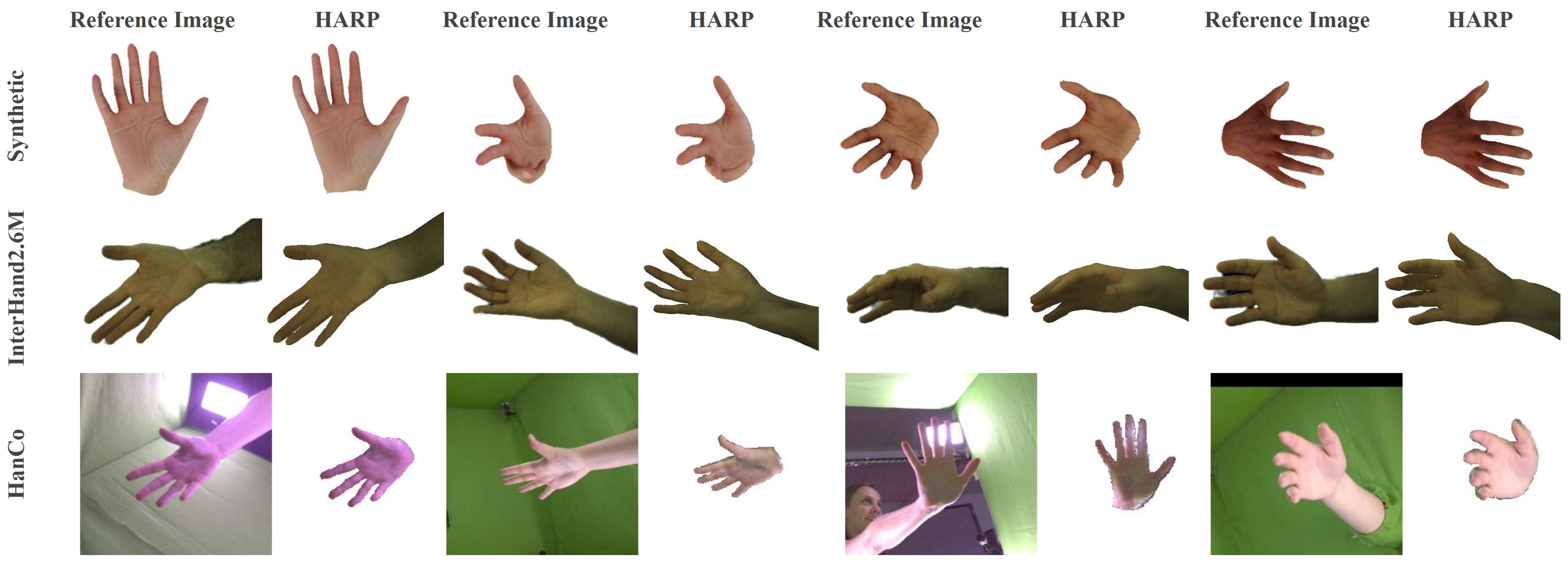}
    \caption{\textbf{Results on various datasets} (our Synthetic, InterHand2.6M \cite{moon2020interhand}, and HanCo \cite{zimmermann21hanco}). HARP is very robust, it can obtain faithful hand avatars on different datasets that have a large variety of capture setups, lighting and image conditions, and textures.}
    \label{fig:other_dataset}
    \vspace{-1em}
\end{figure*}

\myparagraph{Out-of-distribution Hand Appearance.} \label{sec:app_ood}
To demonstrate the advantage of our method to capture out-of-distribution appearance, we compare the optimized results to those of the PCA-based HTML and NIMBLE and the MLP-based NHA on the videos of hands with tattoos, using the same optimization pipeline and objectives.
The results are shown in Tab.~\ref{table:tattoo}.
By not being constrained by the PCA space, we can reasonably capture such out-of-distribution appearance. At the same time, the tattoos are completely discarded by HTML and NIMBLE as they are not in the training set of those models.
The qualitative results Fig.~\ref{fig:tattoo} demonstrate the robustness of our system to capture the non-standard appearance.
Note that we only demonstrate with the tattoo but such appearance deviation can also be scars or nail coloring (please see the Appendix).

\begin{table} 
\footnotesize
\centering
\begin{tabular}{lccc}
\toprule
 & L1 $\downarrow$~ & LPIPS $\downarrow$~ & MS-SSIM $\uparrow$~ \\ \hline

\multicolumn{1}{l|}{HTML \cite{HTML_eccv2020}} & 0.018 & 0.121 & 0.836 \\ 
\multicolumn{1}{l|}{NHA \cite{grassal2021neural}} & 0.017 & 0.131 & 0.891 \\
\multicolumn{1}{l|}{NIMBLE \cite{li2022nimble}} & 0.029 & 0.178 & 0.736 \\
\hline
\multicolumn{1}{l|}{HARP} & \textbf{0.012} & \textbf{0.080} & \textbf{0.897} \\ 
\bottomrule
\end{tabular}
\caption{{Evaluation on out-of-distribution appearance}.}
\label{table:tattoo}
\end{table}

\myparagraph{Differentiable Self-shadow Modeling.}
Self-shadowing between fingers and palm is almost inevitable due to highly articulated and dexterous hand movements. 
However, none of the baselines can properly capture and model self-shadowing.
To validate the effectiveness of our shadow-aware differentiable shader, we compare the appearance reconstruction quality with and without shadow modeling.
The quantitative comparison is shown in Tab.~\ref{table:shadow_comparison}.
Without properly modeling shadow, the optimized color at each pixel averages the color when that pixel is in and out of the shadow, resulting in dark patches baked into the texture.

\begin{table} 
\footnotesize
\centering
\begin{tabular}{lccc}
\toprule
  & L1 $\downarrow$~ & LPIPS $\downarrow$~ & MS-SSIM $\uparrow$~ \\ \hline

\multicolumn{1}{l|}{HARP w/o shadow} & 0.0129 & 0.054 & 0.940 \\
\multicolumn{1}{l|}{HARP w/ shadow} & \textbf{0.0123} & \textbf{0.051} & \textbf{0.943} \\ 
\bottomrule
\end{tabular}
\caption{Comparison between HARP with and without self-shadow modeling on the shadow portion of our captured dataset. Qualitative comparison in the Appendix.}
\label{table:shadow_comparison}
\end{table}

\subsection{Hand Pose Reconstruction via Appearance Modeling} \label{sec:pose_eval}

With a realistic hand appearance obtained from a video using HARP, we demonstrate that such appearance information can improve the RGB hand pose estimation from the same identity using differentiable rendering if the appearance is known.
Intuitively, with differentiable rendering, the optimization should be able to improve the initial prediction using only the hand mask to refine the pose.
However, we observe that such masks are not always useful depending on the poses and viewpoints (see Appendix). In such cases, knowing the hand appearance supplements the mask in guiding the optimization toward the correct pose.

To the best of our knowledge, we are the first to qualitatively and quantitatively demonstrate such improvement for hand pose estimation.
We believe that the main component for enabling such improvement lies in the design of the rendering process such as shading and visibility check. 
However, such topics did not receive much attention in the hand community especially in the context of pose estimation prior to this work.
We analyze the scenario in which such improvement is possible in the Appendix.

In Tab.~\ref{table:pose_estimation}, we compare the hand pose error between METRO \cite{lin2021metro}, HARP with silhouette loss only (HARP-sil), a normal HARP (HARP-full), and HARP with a known appearance that is frozen during optimization (HARP-known).
The known appearance is obtained from running HARP on a simple video of the same identity.
The results indicate that HARP can leverage the differentiable rendering of appearance to improve poses.
When the appearance is known in advance (HARP-known), the optimization can perform better
as it avoids the shadow and lighting effect that will be baked into the texture during the optimization.

\begin{table} 
\footnotesize
\centering
\begin{tabular}{l|c|ccc}
\toprule
 Dataset & METRO & HARP-sil & HARP-full & HARP-known  \\ \hline

Synthetic & 6.12 & 6.16 & 6.04 & \textbf{5.65} \\ 

\bottomrule
\end{tabular}
\caption{Pose error on synthetic sequences (PA-MPVPE in mm).}
\label{table:pose_estimation}
\vspace{-1em}
\end{table}

\subsection{Novel View and Pose Synthesis} \label{sec:app_novel}

The explicit geometry obtained using HARP also enables consistent 3D geometry and appearance rendering across different poses and viewpoints. We evaluate this feature by rendering the known appearance onto the hands in novel views and novel poses, using the test sequences of our dataset.
To match the pose in the test sequences, we optimize only the pose parameter $\gamma$ (see Sec.~\ref{sec:hand_rep}) with respect to the hand masks while freezing all other components.
The quantitative results shown in Tab.~\ref{table:result_texture_syn} suggest that HARP can produce consistent appearances in novel views and poses.

\begin{table} 
\footnotesize
\centering
\begin{tabular}{lcccc}
\toprule
 & IoU $\uparrow$~ & L1 $\downarrow$~ & LPIPS $\downarrow$~ & MS-SSIM $\uparrow$~ \\ \hline

\multicolumn{1}{l|}{METRO \cite{lin2021metro}} & 0.651 & - & - & - \\ 
\multicolumn{1}{l|}{S2Hand \cite{chen2021s2hand}}  & 0.448 &  0.059 & 0.271  & 0.598  \\ 
\multicolumn{1}{l|}{HTML \cite{HTML_eccv2020}}  & 0.754 & 0.040 & 0.144 & 0.771\\
\multicolumn{1}{l|}{NHA \cite{grassal2021neural}}  & 0.826 &  0.034 & 0.134  & \textbf{0.850}  \\ 
\multicolumn{1}{l|}{NIMBLE \cite{li2022nimble}}  & 0.752 &  0.038 & 0.157  & 0.789  \\ 
\hline
\multicolumn{1}{l|}{HARP} & \textbf{0.870} & \textbf{0.029} & \textbf{0.105} & 0.831 \\ 
\bottomrule
\end{tabular}
\caption{Appearance synthesis evaluation on the test sequences.}
\vspace{-2em}
\label{table:result_texture_syn}
\end{table}

\section{Conclusion and Limitation}
In conclusion, we present \methodname, a method for reconstructing personalized hand geometry and appearance from monocular RGB sequences.
Starting from a parametric hand model as a geometry backbone, HARP refines the surface to a personalized hand shape. The texture is factorized into an albedo and a normal map. 
The resulting hand model is robust when rendered in novel views and novel poses, outperforming existing baselines both qualitatively and quantitatively. 
Furthermore, HARP is efficient, scalable, and compatible with traditional rendering pipelines. It provides a foundation for the realistic experience of personalized hands in AR/VR applications. 

\myparagraph{Limitation.} 
As our system assumes only a single light source and ambient light, with no specular effect, its ability to replicate the appearance under other lighting assumptions might still be limited. Incorporating an environment map, modeling bounced light, as well as increasing the resolution of the rendered texture are all interesting steps for future works toward a more photorealistic rendering.

\myparagraph{Acknowledgement.}
We sincerely acknowledge Shaofei Wang and Marko Mihajlovic for the discussions, and 
Malte Prinzler for the help with the Neural Head Avatar baseline.
This work was supported by the SNF grant 200021 204840 and an
ETH Z{\"u}rich Postdoctoral Fellowship.


{\small
\bibliographystyle{ieee_fullname}
\bibliography{reference}
}

\newpage
\clearpage

\begingroup
\onecolumn 

\appendix
\begin{center}
\Large{\bf HARP: Personalized Hand Reconstruction from Monocular RGB Videos\\ **Appendix**}
\end{center}

\counterwithin{table}{section}
\counterwithin{figure}{section}
\setcounter{page}{1}

\section{Datasets}
In this section, we describe the details of each dataset used in the experiments.
We reiterate that our goal is to propose a scalable and robust system that can create faithful hand avatars given a short video sequence that is captured by commodity hardwares, such as a smartphone. Such setup facilitates the utility of our method in downstream applications, e.g., personalized hand avatar creation for end-users of AR/VR devices. 
Unfortunately, there is no existing dataset designed and captured for this scenario. Therefore we capture the \textbf{Hand Appearance Dataset} with a smartphone in a room with common lighting conditions (e.g.,~light bulbs on the ceiling). 

To demonstrate that HARP is robust to different capture setups, we additionally test HARP on sequences from the \textbf{InterHand2.6M \cite{moon2020interhand}} and \textbf{HanCo \cite{zimmermann21hanco}} datasets.
To supplement the lack of {\it accurate} ground truth to evaluate the pose refinement results, we create our \textbf{Synthetic Dataset} using a ray tracing engine.
The details of each dataset are as follows:

\myparagraph{Hand Appearance Dataset.}
The dataset is partitioned into three parts as described in the main paper.
All of the sequences are captured with a hand-held smartphone camera in different conditions.
The foreground masks, which include both the hand and arm, are obtained using an off-the-shelf segmentation tool Unscreen \cite{unscreen}.
The images are resized to 448x448 pixels, which we use as a default size in all of our experiments unless indicated otherwise.

\myparagraph{InterHand2.6M \cite{moon2020interhand}.}
We demonstrate HARP's ability to create an avatar from existing datasets on sequences from the Interhand2.6M dataset.
The sequences in the dataset are captured in a capture dome with a multi-view camera rig and uniform lighting.
As the dataset does not include segmentation masks, we obtain foreground-background masks using RVM \cite{rvm_seg}.
We notice that flares from light bulbs in the capture dome often interfere with the segmentation and are sometimes categorized as foreground. We note that such artifact is difficult to remove and could degrade the optimized appearance.
To avoid such artifacts, 
we use 500 frames from \textit{cam400266} of \textit{Capture0/ROM03\_RT\_No\_Occlusion} from the 30-FPS test set (frame 500th to 999th ).
The images are cropped to 334x334 pixels with the mean of hand vertices at the center.

\myparagraph{Synthetic Dataset.}
In order to evaluate the pose refinement results, we create the synthetic dataset with perfect ground truth pose annotations.
The images are rendered using the ray tracing engine Cycles in Blender \cite{blender}. 
We leverage the NIMBLE model \cite{li2022nimble} to obtain the hand meshes and appearances.
The appearances are manually selected to ensure diversity in size and skin color from the appearances sampled from NIMBLE.
Due to the dependency on NIMBLE, the generated hands are truncated at the wrist and the arm is not visible in the images.
For rendering, we use the same Blender settings as the one provided in the demo of NIMBLE.
The hand motions are the same in all of the generated sequences. The main differences between each sequence are the viewpoint, the hand shape, and the hand appearance.
For each identity, we generate two motions, one is a hand-flipping motion and another one with finger movement.
The motions are 5 seconds long at 30 FPS.
Fig.~\ref{fig:ablation_synthetic_data} shows the images from our Synthetic dataset.

\begin{figure}[b]
    \centering
    \includegraphics[width=0.7\linewidth]{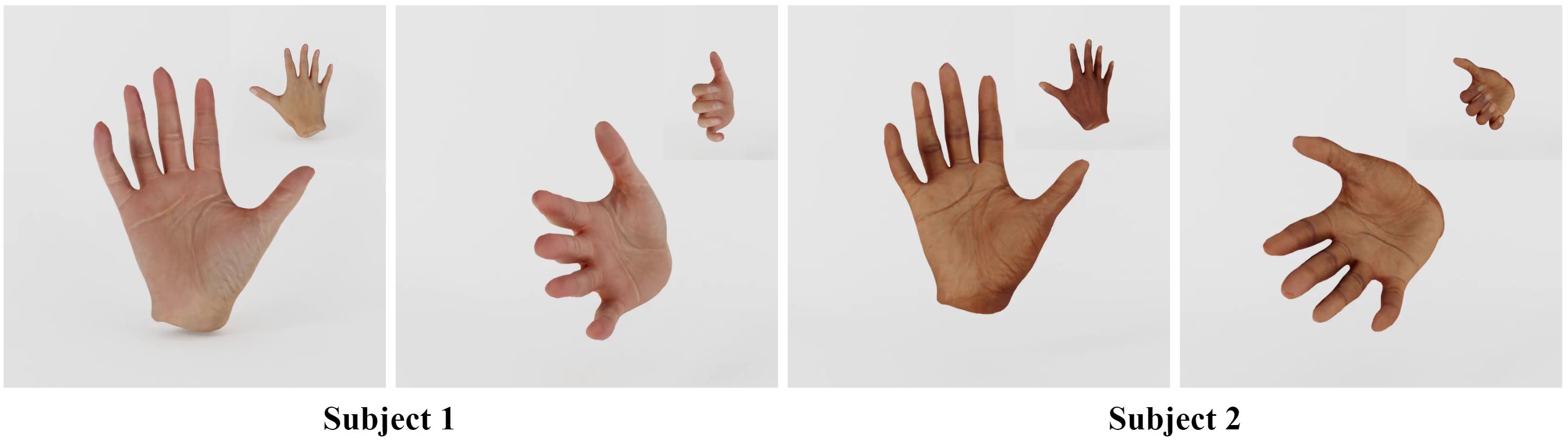}
    \caption{\textbf{Synthetic dataset}. Each image shows the first frame and the middle frame from each sequence.}
    \label{fig:ablation_synthetic_data}
\end{figure}

\begin{figure}[]
    \centering
    \includegraphics[width=0.9\linewidth]{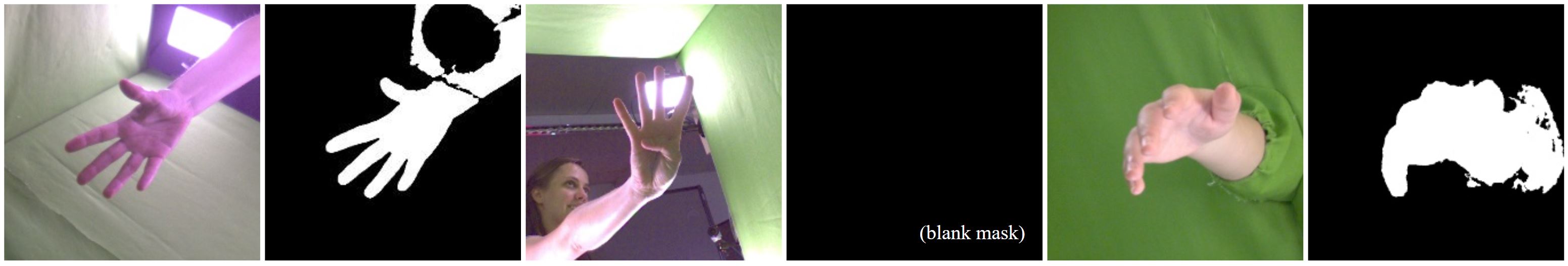}
    \caption{\textbf{Sample images and masks from the HanCo \cite{zimmermann21hanco} dataset}. The dataset is not suitable for avatar creation because of the low resolution and unrealistic light stage.}
    \label{fig:ablation_hanco_data}
\end{figure} 

\myparagraph{HanCo \cite{zimmermann21hanco}.}
We show the results from the following sequences in Fig.~6 in the main paper: 0 (cam7), 2 (cam1), 10 (cam4), and 27 (cam4).
Both sides of the hand are visible in these sequences.
The provided foreground-background masks are used as input for the optimization.
Nevertheless, note that the dataset is not suitable for avatar creation and should be treated only as a reference due to the low resolution of the image at 224x224 pixels and the unrealistic light stage. Samples images are shown in Fig.~\ref{fig:ablation_hanco_data}.

\section{Baselines}
To ensure a fair comparison between HARP and the baseline methods, all of the baselines are optimized at test time, with an equivalent number of epochs when possible.
We use the officially released code from each baseline. 
As some of the baselines are not designed for hand avatar creation via optimization, we need to make adjustments and modifications, which we describe for each baseline in the followings.
In the case of HTML \cite{HTML_eccv2020} and NIMBLE \cite{li2022nimble}, we use the same optimization pipeline as in HARP, replacing only the relevant parts with their models.
All the methods take the same images and pose initialization as input.

\myparagraph{HTML \cite{HTML_eccv2020}.} 
We replace the UV texture and normal map of HARP with the texture produced by the HTML model. The HTML texture vector is optimized instead of the HARP texture.
As the HTML texture is defined on the surface of MANO \cite{MANO2017} hand mesh, we use the MANO template instead of our HARP template.
In addition, we allow vertex displacement along normals in the same manner as in HARP.
The same optimizers and loss terms are used in the optimization.

\myparagraph{NIMBLE \cite{li2022nimble}.}
As the NIMBLE model provides both shape and appearance space, we replace the HARP hand geometry and appearance with NIMBLE. The NIMBLE pose, shape, and appearance parameters are updated during the optimization. The same optimizers and loss terms are applied.
For initialization, we fit NIMBLE to the METRO prediction instead of our template, using the same formulation as described in Sec.~\ref{sup_sec:init}.

\myparagraph{S2Hand \cite{chen2021s2hand}.}
We modified the original S2hand code which predicts the appearance from the input image to allow the optimization with photometric losses used in HARP.
Because the S2Hand model requires ground truth camera parameters for projecting the predicted mesh onto the image frame, during evaluation we optimize the camera parameters by minimizing the photometric loss with respect to the input image. This optimization is done separately to obtain the best match for each frame.
We acknowledge that the optimized texture quality might be affected by the less accurate pose estimation from S2Hand. Nevertheless, due to the fact that S2hand requires ground truth camera intrinsics for image projection but METRO estimates camera extrinsics for a fixed set of intrinsics, it is not possible to optimize S2Hand with our coarse initialization.

\myparagraph{Neural Head Avatar \cite{grassal2021neural}.}
As this method is designed specifically for reconstructing a human head avatar, we make several modifications to adapt it to the hand avatar creation task.
Notably, we drop the dependency on face segmentation, landmark, and predicted normals from the model.
For segmentation, there are only foreground and background, which are the same as the ones used in other baselines.
For landmarks, the landmark locations are replaced with hand key point locations.
The predicted normal input is discarded from the model. In terms of implementation, a fixed identity is used in place of the unavailable input.
For the geometry, we replace the FLAME \cite{FLAME:SiggraphAsia2017} model with our hand template with arm (details in Sec.~\ref{sup_sec:template}).
We follow the official code and instructions for training and evaluating the results.

\section{Implementation Details}

\subsection{Hand Templates} \label{sup_sec:template}
In order to create a hand avatar, we observe that the truncation at the wrist in the MANO \cite{MANO2017} model is problematic to the appearance-creation process and does not reflect the reality that a hand is always attached to an arm. Thus, we implement a version of the hand model with an arm, which we derive from SMPLX \cite{SMPL_X2019} by truncating the mesh at the elbow, moving the root joint to the right-hand wrist, and linearly subdividing the mesh once.
As a result, our hand model has two modes: hand-only and hand-with-arm, which can be used interchangeably depending on the available mask.
The comparison between the template meshes is shown in Fig.~\ref{fig:hand_template}.

\begin{figure}[h]
    \centering
    \includegraphics[width=0.55\linewidth]{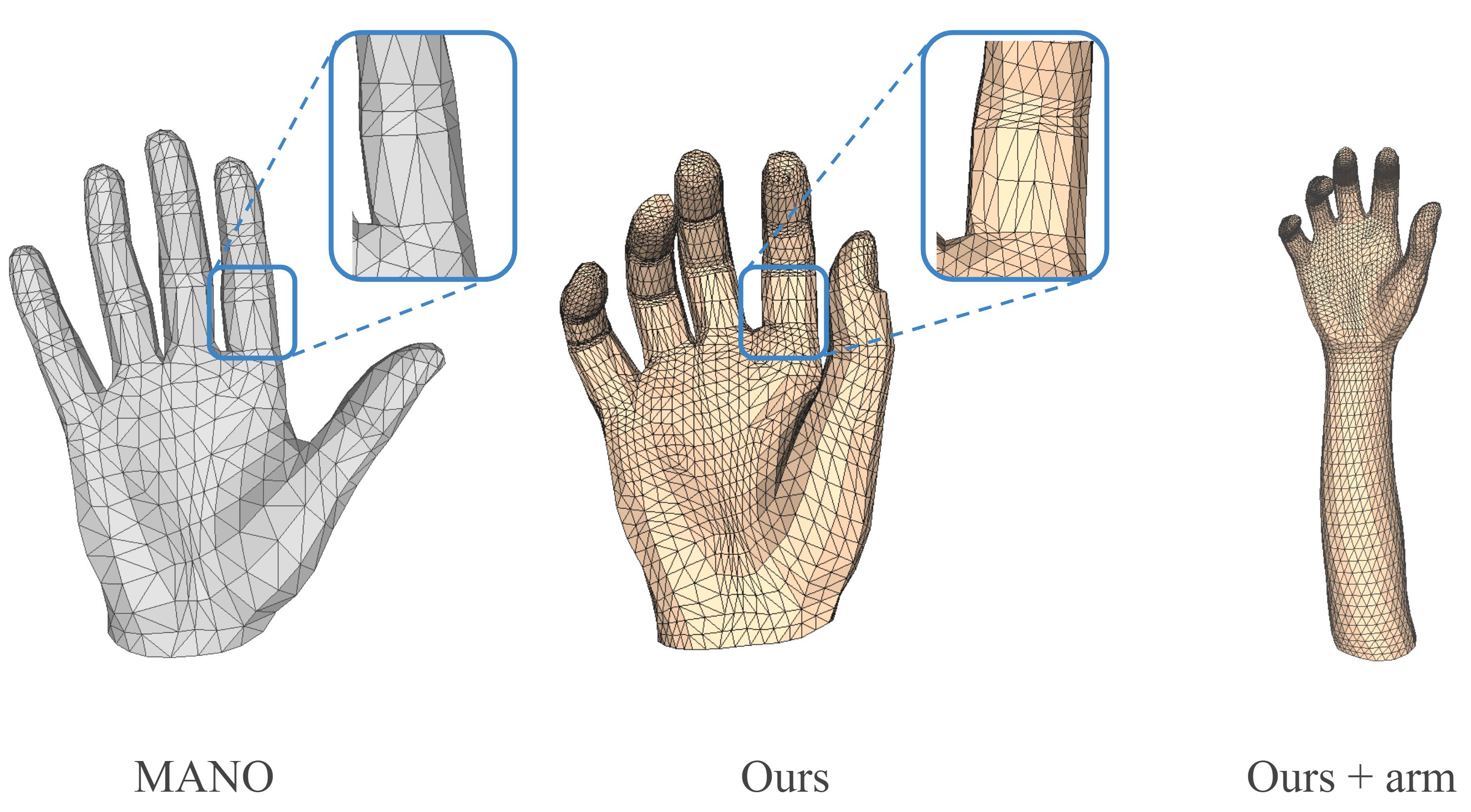}
    \caption{ Template Meshes.}
    \label{fig:hand_template}
\end{figure}

\subsection{Initialization} \label{sup_sec:init}
\myparagraph{Mesh Initialization.} As discussed in the main paper, before we start the optimization process, the hand pose and shape need to be initialized. Therefore, we employ the pose estimator METRO \cite{lin2021metro} which estimates the camera translations with fixed intrinsic parameters. We average the camera translations across all frames from the same sequence and fix them throughout the optimization.
However, because the METRO model predicts the mesh vertex locations directly, the predicted meshes cannot be used for animation. To this end, we fit the hand template to METRO predictions by optimizing pose parameters $\gamma$ and shape parameters $\beta$ using the following energy term $E$:

\begin{align*}
    E &= \sum_v^V \norm{v_t - v_p}^2 \\
    \text{where~} V_t &=  \mathcal{M}(\gamma, \beta)\\
\end{align*}
where $V$ is the set of MANO vertices in the template, and $v_p$ is the predicted location from METRO.
Note that this optimization is possible because METRO prediction and MANO model share the same template mesh.
To avoid local minima during fitting, we re-run the optimization process if the mean distance between METRO vertices and the optimized vertices is more than 1 cm.

\subsection{Optimization}
From the initialization, we first optimize using only the geometry term $E_{geo}$ to reconstruct the hand surface. We then jointly optimize both the geometry and the appearance with the addition of $E_{app}$.
Once the geometry is stable in the joint optimization, we then freeze the geometry optimization and continue to refine the appearance with only $E_{app}$.
Concretely, to obtain the personalized hand pose, shape, and appearance, 
we employ a multi-stage optimization scheme
as follows: (1) geometry optimization, (2) both geometry and appearance optimization, and (3) only appearance optimization. We use the Adam \cite{kingma2014adam} for both $E_{geo}$ and $E_{app}$ optimization.
In total, the optimization takes an average of 80 minutes on a single Nvidia 3090 GPU.

\myparagraph{Geometry Optimization.} 
Given the masked images, we first optimize the pose $\gamma$, shape $\beta$, vertex displacements $D$, translations, and rotations, with respect to the geometry objective $E_{geo}$. 
In this stage, only the geometry energy term $E_{geo}$ is used. loss is used.
We optimize using a learning rate of $1e^{-3}$ for 100 epochs. 

\myparagraph{Joint Optimization.}
After the coarse geometry alignment, we begin the appearance optimization with respect to the appearance objective $E_{app}$. In this stage, both the geometry objective $E_{geo}$ and appearance objective $E_{app}$ are optimized together for 50 epochs to correct geometry misalignment using appearance information on the input images.
We use the learning rate of $1e^{-2}$ for the appearance optimizer.

\myparagraph{Appearance Optimization.} 
Lastly, we refine the appearance with only the appearance objective $E_{app}$ for another 50 epochs. This step focuses on retrieving fine texture details which are difficult to optimize while the geometry is still changing.

\myparagraph{Appearance Regularization Terms.}
To regularize the reconstruction of the UV texture and normal map, we define the appearance regularization term in the UV space. 
Let $\mathcal{T}$ be an albedo map and $\mathcal{G}$ be a normal map, and $I$ is a pixel in the UV space:

 \begin{align}
    E_{app\_reg} &= E_{t\_reg} + E_{n\_reg}, \\
    E_{t\_reg} &= \sum_I \frac{1}{3} \norm{ \mathcal{T}(I) - \mathcal{T}(I + \epsilon_1)}_1, \\
    E_{n\_reg} &= \sum_I \frac{1}{3} (\norm{ \mathcal{G}(I) - \mathcal{G}(I + \epsilon_2)}_1 + \norm{ \mathcal{G}(I) - u_z }^2_2 ),
 \end{align}
where $\epsilon_1,\epsilon_2$ are random pixel-space displacements sampled from a Gaussian with a standard deviation of 2, $u_z$ is a unit vector pointing along $z$ direction.
Both terms ensure a smooth transition in the UV space, while the $E_{n\_reg}$ encourages the normal to be close to the surface normal.

\myparagraph{Losses.}
The weights for each energy term are as defined in the table \ref{table:loss}:
 
\begin{table}[h]
\small
\centering
\begin{tabular}{l|l}
    $E_{sil}$ & 7.0 \\
    $E_{init}$ & 10.0 \\
    $E_{verts}$ & 2.0 \\
    $E_{lap}$ & 4.0 \\
    $E_{norm}$ & 0.1 \\
    $E_{arap}$ & 0.2 \\
    $E_{photo}$ & 1.0 \\
    $E_{vgg}$ & 1.0 \\
    $E_{t\_reg}$ & 2.0 \\
    $E_{n\_reg}$ & 0.5
\end{tabular}
\caption{\textbf{Weights for each energy term.}}
\label{table:loss}
\end{table}

\subsection{Lighting Contribution}

As we assume that the hand surface is largely non-reflective, we ignore the specular contribution in our lighting formulation.
In our method, the color at each surface point is only affected by the ambient contribution, which dictates how bright each point is regardless of its position, and the diffuse contribution, which determines the brightness based on the angle between the point normal and the light direction. Diffuse lighting is also affected by the visibility term $V$ that determines direct occlusion with respect to the light source.
A higher ambient contribution will make the shadow less visible and the brightness more uniform.
Figure \ref{fig:light_decomposition} shows the decomposition of each component in our pipeline.
We show the differences between the final albedo after optimizing with and without considering the visibility term $V$ in Fig.~\ref{fig:light_shadow}.

In our experiments, we empirically disable the self-shadowing term of HARP when we compare it with the baselines on the first part of our hand appearance dataset, where there is no hard shadow and dominant light source.
In other experiments, the self-shadowing term is enabled and the ratio between the ambient and diffuse contributions is optimized together with other parameters as described in the main paper.

\begin{figure}[h]
    \centering
    \includegraphics[width=0.8\linewidth]{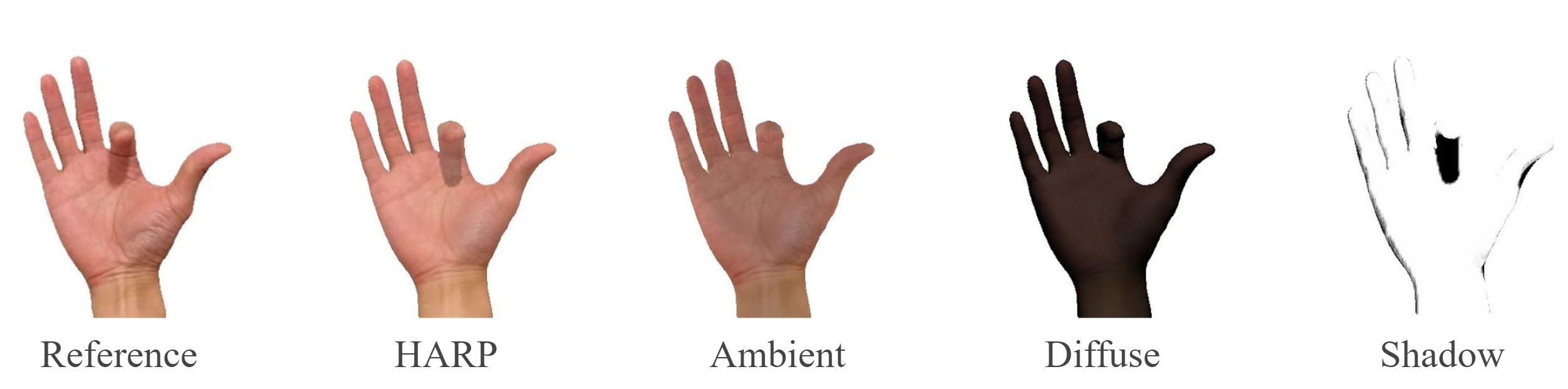}
    \caption{\textbf{Decomposition of color contributions in our rendering pipeline.} }
    \label{fig:light_decomposition}
\end{figure}

\begin{figure}[h]
    \centering
    \includegraphics[width=0.8\linewidth]{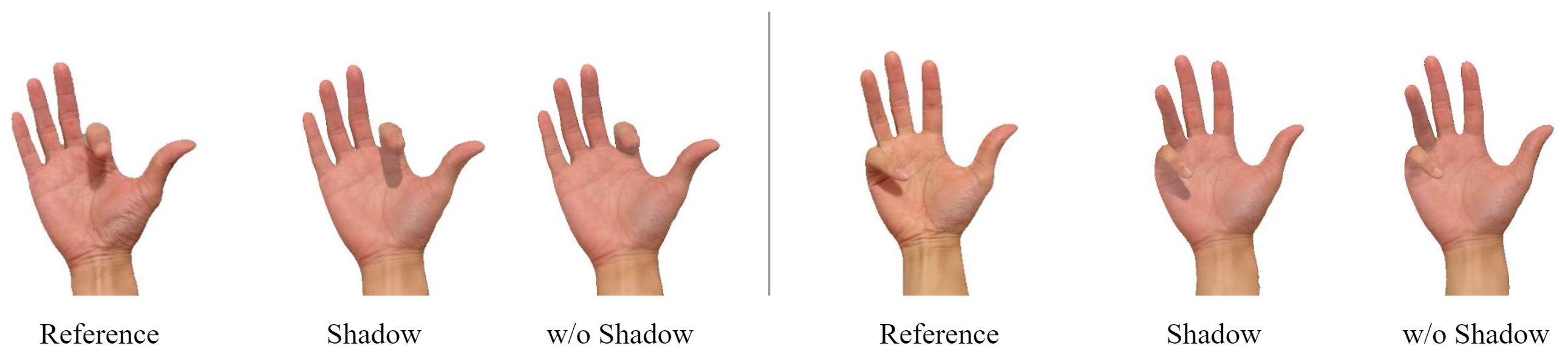}
    \caption{\textbf{Comparison between optimization without and with self-shadowing effect.} Without considering self-shadow, the input images cannot be faithfully reconstructed by the renderer.}
    \label{fig:light_shadow}
\end{figure}

\section{Ablation}
In this section, we discuss the importance and effect of each energy term in our method.

\begin{table} 
\small
\centering
\begin{tabular}{lccc}
\toprule
  & L1 $\downarrow$~ & LPIPS $\downarrow$~ & MS-SSIM $\uparrow$~ \\ \hline

\multicolumn{1}{l|}{w/o $E_{photo}$} & 0.0171 & 0.0693 & 0.906 \\ 
\multicolumn{1}{l|}{w/o $E_{VGG}$} & 0.0164 & 0.0842 & 0.914 \\ 
\multicolumn{1}{l|}{HARP} & 0.0168 & 0.0712 & 0.908 \\ 
\bottomrule
\end{tabular}
\caption{\textbf{Ablation study on the appearance losses.}}
\label{table:ablation}
\end{table}

\myparagraph{Appearance.} 
We observe that, without the perceptual term $E_{VGG}$, the resulting texture looks overly smooth as the colors are averaged over the pixels that map to a slightly different point on the hand surface.
However, without the photometric L1 term $E_{photo}$, the result might contain noisy artifacts.
The qualitative comparison is shown in Table \ref{table:ablation}.

\myparagraph{Geometry.}
Figure \ref{fig:ablation_geo} shows the qualitative comparison between the results from optimizing without a specific shape regularization term.
The as-rigid-as-possible term $E_{arap}$ prevents sharp edges when the mesh is deformed to fit the silhouette.
The vertex displacement term $E_{verts}$ ensures that the deviation from the shape space of the underlining parametric model is minimal, such that the blendshape from the parametric model is still useful when the mesh is reposed.
The other terms including the normal consistency regularization $E_{norm}$ and the Laplacian regularization $E_{lap}$ encourage the surface to be more smooth and less bumpy.
We note that the effect of each term is less noticeable in the rendered image evaluation as the optimization can counteract the geometry change with a texture change.
However, the regularization terms are necessary to ensure mesh integrity for any downstream application.

\begin{figure}
    \centering
    \includegraphics[width=0.8\linewidth]{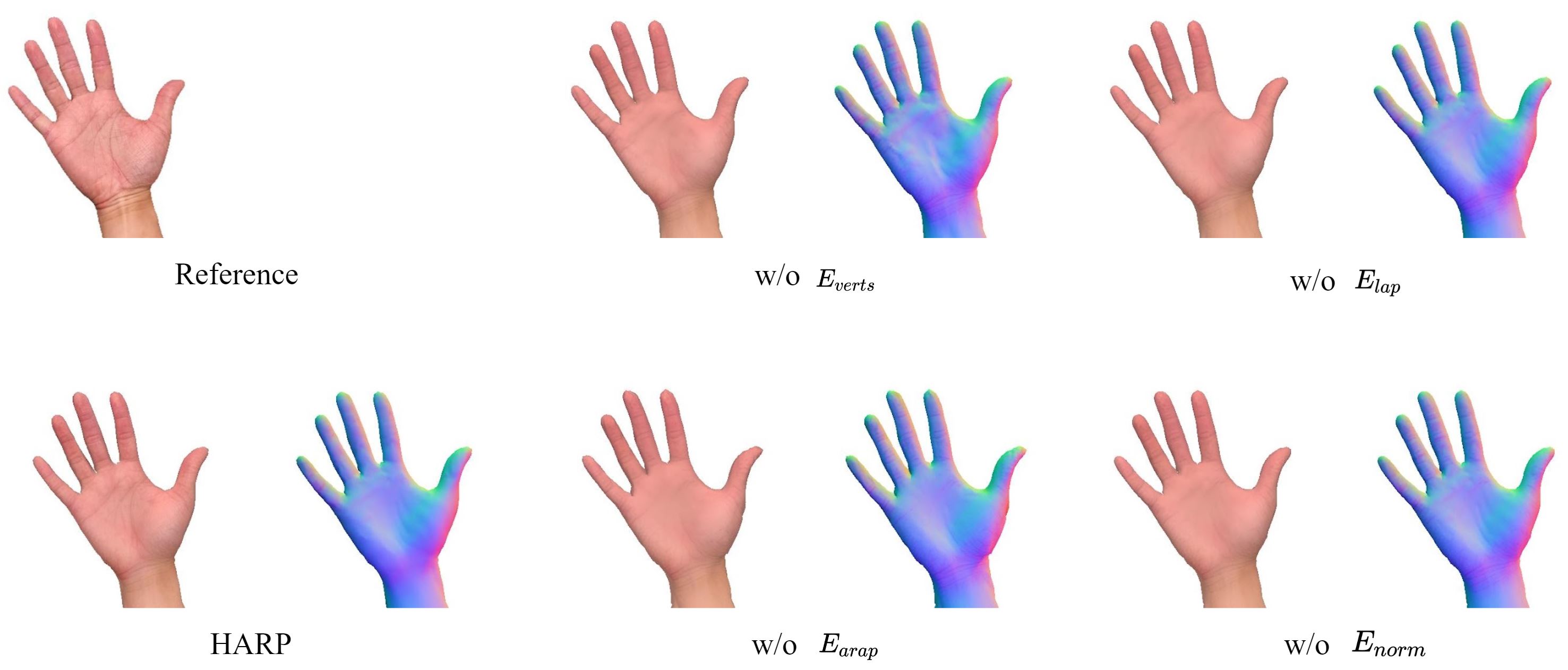}
    \caption{\textbf{Qualitative comparison between optimization results using different geometry regularization terms.}}
    \label{fig:ablation_geo}
\end{figure}

\section{Discussion}

\begin{figure}
    \centering
    \includegraphics[width=0.8\linewidth]{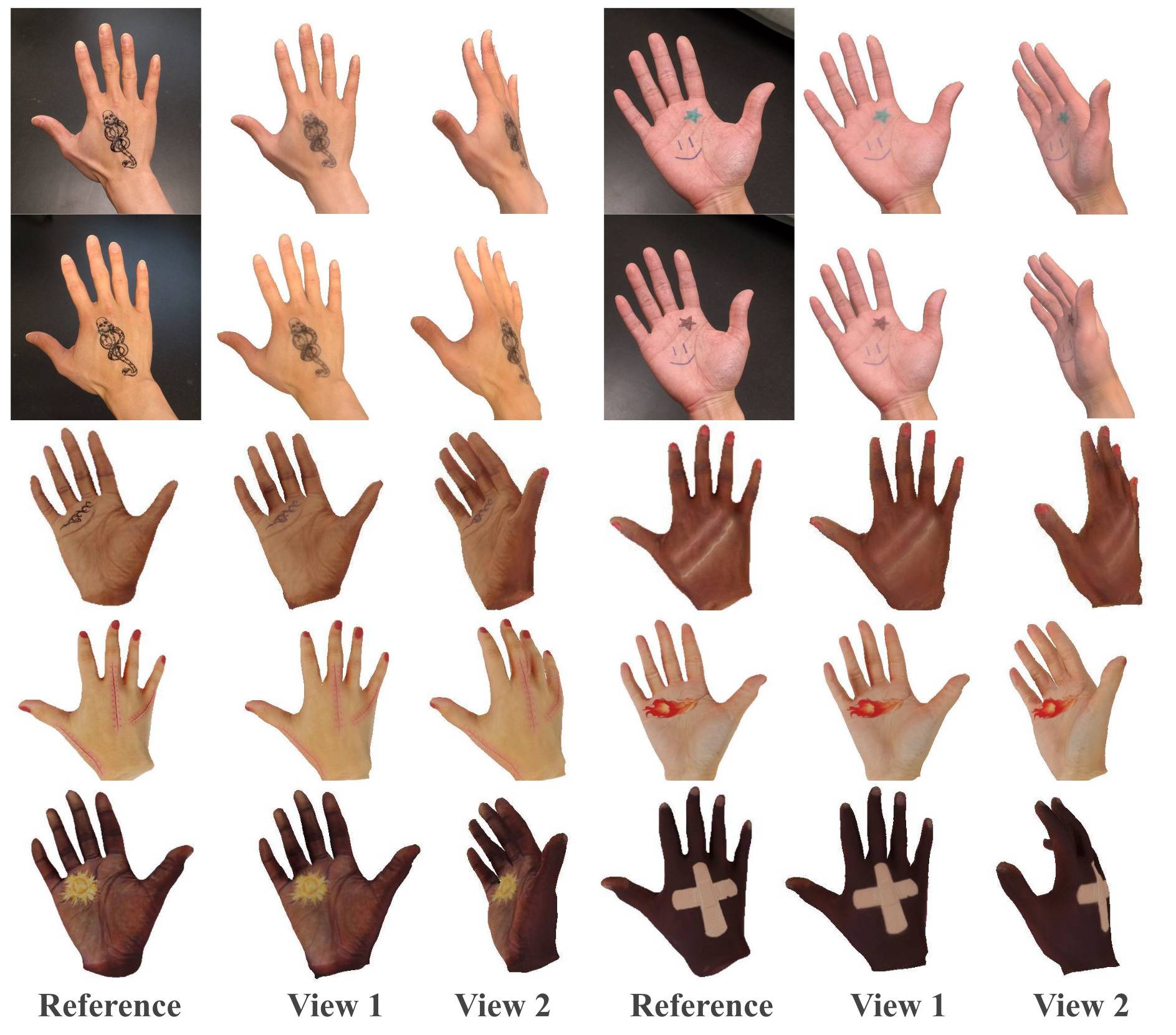}
    \caption{\textbf{Qualitative results of out-of-distribution hands with different lighting conditions.} Our method can accurately capture diverse appearances. 
    The first two rows are captured in different lighting conditions with a different number of ceiling lights. The last three rows are synthetic data rendered with three light sources and ambient light. Please zoom in for details.}
    \label{fig:additional_result}
\end{figure} 

\subsection{Additional Results}
We show additional results in Fig.~\ref{fig:additional_result} that demonstrate our method's ability to capture diverse patterns such as tattoos, nail colors, and scars faithfully on diverse skin colors.

\subsection{Pose-dependent surface deformation}

As our model is built on top of MANO, it has the pose-dependent surface deformation modeled by MANO pose blend shapes. 
Our displacement map is designed to capture detailed, personalized hand shapes that cannot be represented by MANO.
We found that conditioning the displacement map on poses results in two different sets of parameters governing the same surface deformation which could be difficult to optimize.


\subsection{Failure Cases}
In this section, we discuss the noticeable failure cases of our system.
The examples are shown in Fig.~\ref{fig:failure}.
First, HARP mainly uses the silhouette from a monocular view to guide the personalized geometry. As a consequence, it is crucial that the images and the masks provide sufficient information about the shape.
When the data is not sufficient, the hand mesh can deform in an unexpected way to satisfy the mask. 
We show this failure case in Fig.~\ref{fig:failure}(a) where some vertices extend perpendicular to the palm as there are not enough side view images.
Note that the optimization can always compensate for the bumpy geometry with a change in the texture in order to replicate the input images.
A potential solution to this problem is to vary the geometry regularization terms based on the characteristic of the hand in the video.
Second, as our method relies on the initialization from a hand pose estimator and the foreground mask, the final pose and the appearance quality are influenced by the performance of the pose estimator and the  segmentation tool.
In some cases, the pose might be stuck in a local minimum due to the initialization (Fig.~\ref{fig:failure}(b)).

\begin{figure}
    \centering
    \includegraphics[width=0.8\linewidth]{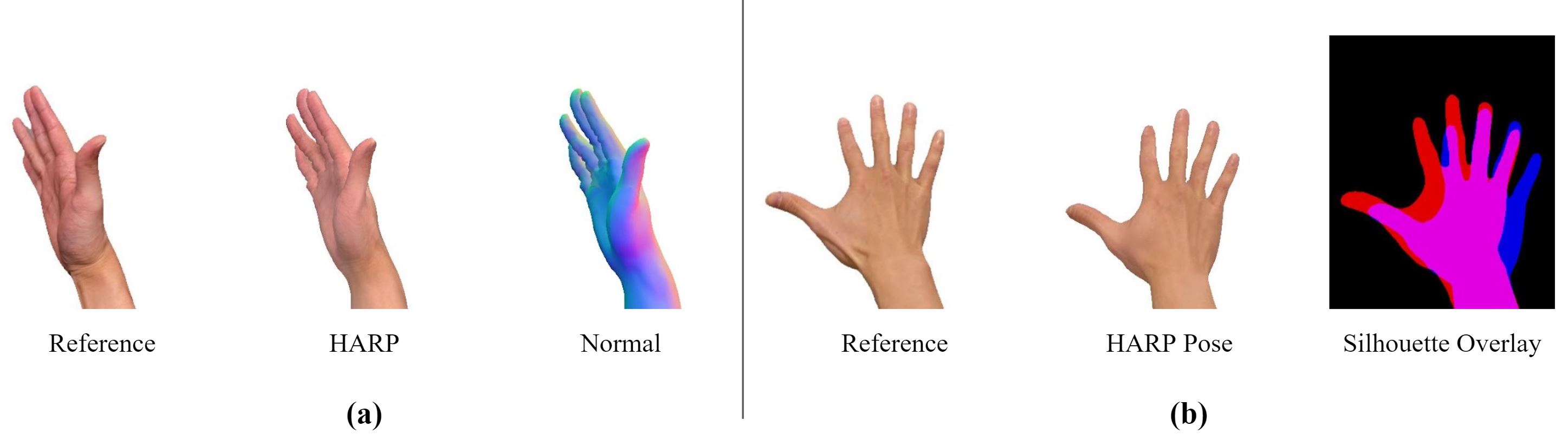}
    \caption{\textbf{Failure cases and limitations.} (a) With limited pixel information, the geometry could deform in an undesirable way as it uses mainly the silhouette for supervision.
    (b) HARP pose optimization is sensitive to initialization and could stuck in local minima when the initialization and foreground mask do not align. In this case, a large immediate increase in silhouette loss will prevent the optimizer from leaving the minima.}
    \label{fig:failure}
\end{figure}

\subsection{Pose Refinement via Appearance Optimization}
In the main paper, we show that by optimizing the hand pose parameters with HARP, we can refine the estimated hand pose to better fit the image, which can lead to a slight improvement in the Procrustes-aligned hand pose error.
Our intuition is that if the hand's appearance is known in advance, it should be possible to leverage pixel color optimization to obtain more accurate poses.
We compare the results between (1) the initial estimation, (2) HARP with only geometry term $E_{geo}$ (HARP-sil), (3) normal HARP (HARP-full), and (4) HARP with known appearance (HARP-known).
\\
\textbf{Case (2)} is a known task that is often associated with a differentiable renderer \cite{ravi2020pytorch3d, liu2019softras} where the silhouette is used to optimize for an object pose.
However, we observe that for a highly articulated object such as a hand, using silhouette alone might not be enough to obtain the correct pose. We visualize such scenarios in Fig.~\ref{fig:mask}.
\\
\textbf{Case (3)} leverages only the appearance consistency within the optimized video. As both the poses and the appearance are optimized together, it is possible to obtain the colors that are associated with wrong poses.
\\
\textbf{Case (4)} leverages the appearance that is obtained from possibly easier hand motion. All of the hand parameters, except for hand poses, are given as initialization. Those parameters, including the appearance, are obtained from running HARP on another sequence.
The given parameters are frozen during the optimization and only the hand poses are updated. All loss terms are the same as normal HARP.

We demonstrate that such pose refinement is possible if the appearance consistency is leveraged in the optimization (both case 3 and case 4).
We acknowledge that our synthetic dataset is small relative to the recent hand pose dataset such as InterHand2.6M\cite{moon2020interhand}
However, due to the lack of ground truth with accurate 3D annotations, we could only perform the experiment on our synthetic dataset where we have \textbf{perfectly accurate ground truth}.
The InterHand2.6M dataset \cite{moon2020interhand}, which offers the hand motions that are the closest to our target use case, reported the MANO ground truth fitting error at around 5 mm \cite{moon2020neuralannot}.
On the other hand, the Procrustes-aligned MANO vertex error of the METRO \cite{lin2021metro} prediction on our selected sequence is at 6 mm.
Any quantitative improvement below 1 mm would be statistically meaningless as it is an order of magnitude smaller than the supposed ground truth error.
Therefore, we do not report the pose refinement on this dataset and other existing datasets due to similar reasons.

\myparagraph{Future work.}
We foresee that the ideal scenario for this use case would be when a user starts using AR/VR equipment, they do a hand-flipping motion to provide a hand appearance. And with that appearance, the pose estimation can be improved.
Practically, however, the optimization speed would still prevent real-time pose refinement. As such improving the speed and pose estimation error which would be interesting to explore in future work.

\begin{figure}
    \centering
    \includegraphics[width=0.7\linewidth]{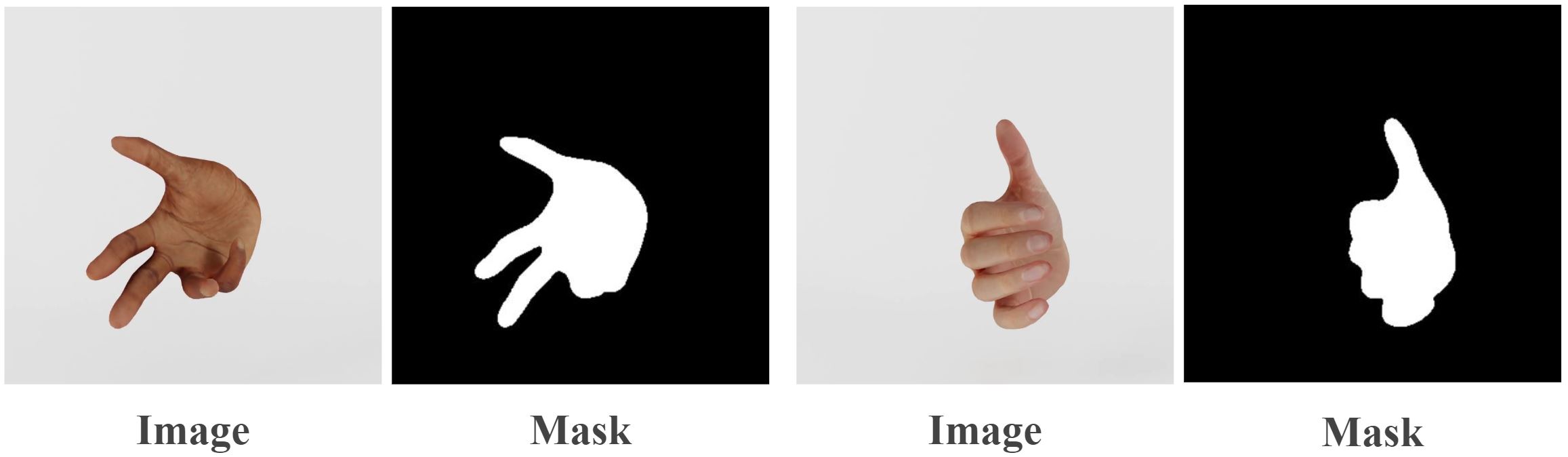}
    \caption{{Example cases where the hand mask is not informative enough for determining the hand pose.}}
    \label{fig:mask}
\end{figure}

\end{document}